\pdfoutput=1
\documentclass{article}

\PassOptionsToPackage{authoryear,round}{natbib}

\usepackage[dblblindworkshop, final]{neurips_2026}
\workshoptitle{Geometric Distributional Deep Learning}

\usepackage[utf8]{inputenc} 
\usepackage[T1]{fontenc}    
\usepackage{hyperref}       
\usepackage{url}            
\usepackage{booktabs}       
\usepackage{amsfonts}       
\usepackage{nicefrac}       
\usepackage{microtype}      
\usepackage{xcolor}         
\usepackage{graphicx}
\usepackage{amsmath}


\newtheorem{theorem}{Theorem}[section]
\newtheorem{proposition}[theorem]{Proposition}

\newenvironment{proofsketch}{\paragraph{Proof.}}{\hfill$\square$}

\newcommand{\SRgrit}{0.090 \pm 0.028}
\newcommand{\MAEgrit}{0.064 \pm 0.009}
\newcommand{\NSgrit}{2.6 \pm 0.5}
\newcommand{\SRMgrit}{0.09}
\newcommand{\ABsrgrit}{2.06 \pm 0.79}
\newcommand{\ABRsrgrit}{1.35 \pm 0.62}
\newcommand{\SRgritdot}{0.532 \pm 0.061}
\newcommand{\MAEgritdot}{0.543 \pm 0.006}
\newcommand{\NSgritdot}{9.4 \pm 1.0}
\newcommand{\ABsrgritdot}{1.15 \pm 0.12}
\newcommand{\ABRsrgritdot}{1.14 \pm 0.11}
\newcommand{\SRdotbias}{0.521 \pm 0.059}
\newcommand{\MAEdotbias}{0.268 \pm 0.068}
\newcommand{\NSdotbias}{9.7 \pm 4.0}
\newcommand{\ABsrdotbias}{1.19 \pm 0.19}
\newcommand{\ABRsrdotbias}{1.20 \pm 0.23}
\newcommand{\SRgps}{0.415 \pm 0.194}
\newcommand{\MAEgps}{0.072 \pm 0.005}
\newcommand{\NSgps}{6.9 \pm 2.2}
\newcommand{\ABsrgps}{1.01 \pm 0.02}
\newcommand{\ABRsrgps}{1.00 \pm 0.01}
\newcommand{\SRgpsnone}{0.507 \pm 0.159}
\newcommand{\MAEgpsnone}{0.200 \pm 0.017}
\newcommand{\NSgpsnone}{8.3 \pm 3.3}
\newcommand{\ABsrgpsnone}{1.11 \pm 0.09}
\newcommand{\ABRsrgpsnone}{2.03 \pm 2.08}
\newcommand{\SRgpsgated}{0.410 \pm 0.119}
\newcommand{\MAEgpsgated}{0.072 \pm 0.007}
\newcommand{\NSgpsgated}{6.0 \pm 1.1}
\newcommand{\ABsrgpsgated}{1.02 \pm 0.02}
\newcommand{\ABRsrgpsgated}{1.00 \pm 0.06}
\newcommand{\SRlappe}{0.511 \pm 0.247}
\newcommand{\MAElappe}{0.661 \pm 0.038}
\newcommand{\NSlappe}{7.8 \pm 3.1}
\newcommand{\ABsrlappe}{1.00 \pm 0.03}
\newcommand{\ABRsrlappe}{1.03 \pm 0.07}
\newcommand{\SRgpsvn}{0.559 \pm 0.177}
\newcommand{\MAEgpsvn}{0.066 \pm 0.015}
\newcommand{\NSgpsvn}{8.8 \pm 1.9}
\newcommand{\ABsrgpsvn}{1.00 \pm 0.02}
\newcommand{\ABRsrgpsvn}{1.00 \pm 0.02}
\newcommand{\SRgpsvnpe}{0.629 \pm 0.199}
\newcommand{\MAEgpsvnpe}{0.086 \pm 0.009}
\newcommand{\NSgpsvnpe}{9.5 \pm 2.2}
\newcommand{\ABsrgpsvnpe}{1.01 \pm 0.03}
\newcommand{\ABRsrgpsvnpe}{1.01 \pm 0.03}
\newcommand{\SRgritvn}{0.095 \pm 0.019}
\newcommand{\MAEgritvn}{0.065 \pm 0.008}
\newcommand{\NSgritvn}{2.7 \pm 0.2}
\newcommand{\ABsrgritvn}{1.93 \pm 1.19}
\newcommand{\ABRsrgritvn}{1.61 \pm 0.18}
\newcommand{\SRgritvnbias}{0.079 \pm 0.027}
\newcommand{\MAEgritvnbias}{0.066 \pm 0.008}
\newcommand{\NSgritvnbias}{2.7 \pm 0.4}
\newcommand{\ABsrgritvnbias}{2.13 \pm 0.87}
\newcommand{\ABRsrgritvnbias}{1.54 \pm 0.50}
\newcommand{\SRgritvnpe}{0.161 \pm 0.039}
\newcommand{\MAEgritvnpe}{0.177 \pm 0.034}
\newcommand{\NSgritvnpe}{3.0 \pm 0.7}
\newcommand{\ABsrgritvnpe}{1.44 \pm 0.31}
\newcommand{\ABRsrgritvnpe}{1.47 \pm 0.39}
\newcommand{\SRentone}{0.131 \pm 0.027}
\newcommand{\MAEentone}{0.082 \pm 0.036}
\newcommand{\NSentone}{3.4 \pm 0.3}
\newcommand{\ABsrentone}{1.83 \pm 0.48}
\newcommand{\ABRsrentone}{1.51 \pm 0.65}
\newcommand{\SRenttwo}{0.127 \pm 0.031}
\newcommand{\MAEenttwo}{0.089 \pm 0.010}
\newcommand{\NSenttwo}{3.3 \pm 0.3}
\newcommand{\ABsrenttwo}{1.71 \pm 0.64}
\newcommand{\ABRsrenttwo}{1.41 \pm 0.37}
\newcommand{\SRdotlogit}{0.680 \pm 0.098}
\newcommand{\MAEdotlogit}{0.373 \pm 0.224}
\newcommand{\NSdotlogit}{13.1 \pm 3.1}
\newcommand{\SRMdotlogit}{0.68}
\newcommand{\ABsrdotlogit}{1.22 \pm 0.43}
\newcommand{\ABRsrdotlogit}{1.23 \pm 0.46}
\newcommand{\SRblind}{0.214 \pm 0.143}
\newcommand{\MAEblind}{0.542 \pm 0.018}
\newcommand{\NSblind}{3.6 \pm 2.0}
\newcommand{\ABsrblind}{1.17 \pm 0.07}
\newcommand{\ABRsrblind}{1.25 \pm 0.17}
\newcommand{\SRnoclamp}{0.113 \pm 0.039}
\newcommand{\MAEnoclamp}{0.063 \pm 0.018}
\newcommand{\NSnoclamp}{3.0 \pm 0.2}
\newcommand{\ABsrnoclamp}{1.91 \pm 1.22}
\newcommand{\ABRsrnoclamp}{1.37 \pm 0.24}
\newcommand{\SRofficial}{0.084 \pm 0.037}
\newcommand{\MAEofficial}{0.059 \pm 0.006}
\newcommand{\NSofficial}{2.5 \pm 1.4}
\newcommand{\ABsrofficial}{1.83 \pm 1.14}
\newcommand{\ABRsrofficial}{1.55 \pm 0.48}
\newcommand{\ABsrdotmin}{1.00}
\newcommand{\ABsrdotmax}{1.22}
\newcommand{\RAYminrest}{1.13}
\newcommand{\RAYmaxrest}{5.45}
\newcommand{\RAYvnpe}{0.71--0.92}
\newcommand{\SATends}{96}
\newcommand{\NIhfmaxgrit}{1.3}
\newcommand{\NIhftwomaxgrit}{0.2}
\newcommand{\NIhfeightmaxgrit}{0.3}
\newcommand{\NIhfmaxnoclamp}{54.0}
\newcommand{\NIhftwomaxnoclamp}{1.4}
\newcommand{\NIhfeightmaxnoclamp}{16.2}
\newcommand{\NIhfmaxofficial}{19.4}
\newcommand{\NIhftwomaxofficial}{0.4}
\newcommand{\NIhfeightmaxofficial}{5.8}
\newcommand{\NIhfmaxdotlogit}{66.9}
\newcommand{\NIhftwomaxdotlogit}{38.8}
\newcommand{\NIhfeightmaxdotlogit}{62.0}
\newcommand{\NIhfmaxgritdot}{51.9}
\newcommand{\NIhftwomaxgritdot}{34.1}
\newcommand{\NIhfeightmaxgritdot}{48.1}
\newcommand{\NIhfmaxgps}{32.1}
\newcommand{\NIhftwomaxgps}{11.5}
\newcommand{\NIhfeightmaxgps}{26.2}
\newcommand{\LMsrA}{0.95--1.00}
\newcommand{\LMsrB}{0.39--1.00}
\newcommand{\CERTmax}{5.1}
\newcommand{\SPREADmax}{8.7\times10^{-5}}
\newcommand{\SPREADdotbias}{0.016}
\newcommand{\ORBVIOLdotbias}{2}
\newcommand{\PROBEdevmax}{7.8\times10^{-7}}
\newcommand{\PROBEdevvn}{2.5\times10^{-4}}
\newcommand{\PROBEdevdotlogit}{0.96}
\newcommand{\PROBEfpdot}{9.6\times10^{-4}}
\newcommand{\RANGEmin}{2.74}
\newcommand{\RANGEmax}{11.69}
\newcommand{\DELTAzinc}{2.24}
\newcommand{\DELTApep}{4.74}

\title{Why don't Graph Transformers attend to the same token?}

\author{Alexandre Bismuth\\BOLD Lab\\University of Oxford
}

\begin{document}

\maketitle

\begin{abstract}
In large language models (LLMs), attention concentrates heavily on the first token. This `attention sink' is believed to protect deep models against over-mixing. Even though graph transformers face a related pressure, we find that a SoTA graph transformer, GRIT, rarely forms sinks on molecules, and that it sinks less when trained on larger graphs, the opposite of the trend in LLMs. In principle, a graph can prevent sinks: we prove and show empirically that its symmetry and locality limit the attention any node can receive. On molecules, however, with distinguishable atoms and global attention, neither limit prevents sinks; what stops them lies in the model. We find that the attention scoring function is the main cause. Replacing GRIT's gated logit with a dot product raises its sink rate from $\SRMgrit$ to $\SRMdotlogit$ on identical graphs. In Graph Transformers using dot-product attention, removing the sink gives about the same error as removing a random node, while GRIT's most-attended nodes matter to its predictions. Attention sinks are therefore a property of dot-product attention rather than of Transformers. In our experiments, they form when dot-product scoring has a distinguishable node to concentrate on.\footnote{The code can be found at \url{https://github.com/alexandre-bismuth/graph-transformer-sinks}.}
\end{abstract}

\section{Introduction}
\label{sec:intro}

A striking empirical property of Large Language Models (LLMs) is the \emph{attention sink} \citep{xiao2024efficient}. In Transformers, an attention sink is a token that receives a large share of all attention: in Llama 3.1 405B, nearly $80\%$ of attention heads sink on the semantically meaningless <BOS> token \citep[sink metric of \citealp{gu2024attention} at threshold $\varepsilon=0.8$;][]{barbero2025llms}. Far from wasting compute, sinks are thought to slow the mixing of information between tokens and so protect deep models against `over-mixing' \citep{overmixingIntro1,overmixingIntro2,overmixingIntro3,barbero2025llms}, and larger models and longer training contexts both bring stronger sinks \citep{barbero2025llms}.

Graph Transformers face a related pressure: depth causes over-smoothing \citep{oversmoothing}, and global attention makes it worse, since softmax over ever more nodes must disperse toward uniform mixing \citep{globalAttention}. A designated sink node might be a natural defence, yet nothing in an ordinary graph plays the role of <BOS>: a graph has no first token, and permutation equivariance forbids preferring one of two nodes the model cannot tell apart. Sinks are documented beyond causally masked LLMs, in bidirectional diffusion language models \citep{rulli2025attention}, as massive activations in attentional GNNs \citep{bini2024massive}, and when LLMs read serialised graphs \citep{guan2025attention}; concurrent work argues that graphs lack natural sinks \citep{siggate2026}. What is missing is an account of \emph{where} a sink can form on a graph at all, and of \emph{which} architectural ingredient decides whether one does.

\begin{figure}[t]
    \centering
    \includegraphics[width=\textwidth]{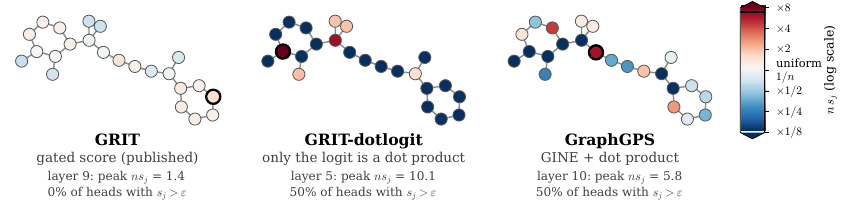}
    \caption{Changing only the scoring function creates a sink. Each atom of a $23$-atom ZINC test molecule is shaded by the attention it receives, averaged over queries and heads at each model's peak-sinking layer: blue below the uniform share $1/23$, red above. Swapping only GRIT's gated logit for a dot product (middle) moves attention off the other atoms and onto one, as in GraphGPS (right).}
    \label{fig:headline}
\end{figure}

We ask whether Graph Transformers form attention sinks and, if not, what prevents them. We study GRIT \citep{GRIT}, a state-of-the-art Graph Transformer without message passing, so attention alone mixes information between nodes. We find that GRIT rarely forms sinks on molecules, and sinks even less when trained on larger graphs (Section~\ref{sec:nosinks}), but that swapping in dot-product attention makes it sink far more often (Section~\ref{sec:scoring}). We investigate two suspects. The graph itself is not the barrier: we prove three bounds on received attention, from symmetry, locality, and the spread of attention scores, and find both graph-side bounds loose on real molecules (Section~\ref{sec:theory}). The attention scoring function is the main cause, as the logit swap of Figure~\ref{fig:headline} and accuracy-matched controls show (Section~\ref{sec:scoring}).

\section{GRIT rarely forms attention sinks on molecules}
\label{sec:nosinks}

Graph structure enters GRIT \citep{GRIT} through Relative Random Walk Probabilities (RRWP), which initialise a learned pair representation. GRIT scores a pair by gating the sum of query and key with that representation, passing it through a signed square root, and clipping it to $[0,5]$ before a scalar readout; dot-product attention instead takes the inner product of query and key, which can grow with the product of the two node norms:
\begin{equation}\label{eq:score}
z^{\mathrm{GRIT}}_{uj}=a^\top\min\bigl\{5,\operatorname{ReLU}\bigl(\operatorname{ssq}((q_u+k_j)\odot w_{uj})+b_{uj}\bigr)\bigr\},\qquad z^{\mathrm{dot}}_{uj}=q_u^\top k_j/\sqrt{h},
\end{equation}
with $\operatorname{ssq}(x)=\operatorname{sgn}(x)\sqrt{|x|}$, $w_{uj},b_{uj}$ linear in the pair representation, and $h$ the head dimension. Clipping the scalar logit instead, as the official code does, gives the same sink rate (Appendix~\ref{apd:figures}). We compare against GraphGPS \citep{GraphGPS} with RWSE \citep{RWSE}, a hybrid of message passing and dot-product attention. We train on ZINC-12k, a regression task on graphs of roughly 23 nodes, and on Peptides-func \citep{LRGB}, a classification task on larger graphs, both whole and split by node count into four quartiles (mean sizes 56 to 245), each trained on 2{,}023 graphs at batch size 16 to remove the dataset-size confound. Our results are means with a $95\%$ CI on the full test set over three seeds (five for GraphGPS and the quartiles).

Our central measurement is the \textbf{sink score} of a node, the attention it receives averaged over all queries \citep{barbero2025llms,gu2024attention}. Following \citet{gu2024attention}, a node is a sink in a head when its sink score exceeds $\varepsilon = 0.3$, about seven times the uniform share on a 23-node graph. The \textbf{sink rate} is the largest fraction of heads that share one sink node, averaged over test graphs, at the peak-sinking layer.

On ZINC, GRIT's sink rate is $\SRgrit$: fewer than one head in ten sinks on the same node, against $\SRgps$ for GraphGPS and \LMsrA\ for small language models at the same context length. GRIT stays below GraphGPS at every threshold from $0.1$ to $0.8$ (Appendix~\ref{apd:figures}). The trend with size also runs backwards. Language models trained on longer contexts form stronger sinks \citep{barbero2025llms}, whereas GRIT's sink rate falls from $0.238$ on the smallest Peptides-func quartile to $0.045$ on the largest. As in language models, where sink tokens develop unusually large norms \citep{queipo2025attention}, GRIT's largest squared node norm grows across layers from $2.4$ to $4.8$ times the squared mean norm. Nonetheless, sinks stay rare, so something in GRIT keeps large norms from attracting attention (Section~\ref{sec:scoring}).

\section{What can prevent a sink?}
\label{sec:theory}

Three things can prevent a node from accumulating attention from many others at once: symmetry, since an equivariant model must treat nodes the labelled graph does not distinguish alike; locality, since nodes that attend mostly to their neighbourhood leave little attention for distant ones; and scoring range, since softmax concentrates only when logits spread far apart.

\textbf{Notation.} We write $A\in\mathbb{R}^{n\times n}$ for the attention matrix of a single layer and head, with logits $z_{uv}$, and define the sink score as $s_j(A)=\tfrac1n\sum_{u}A_{uj}$. We assume permutation equivariance, i.e.\ $A(PZ)=PA(Z)P^\top$ for every permutation matrix $P$ and input $Z$ containing the graph, node features and positional encodings. Let $\mathcal{O}(j)$ be the orbit of $j$ under the automorphisms of $Z$ (the nodes equivariant models must treat like $j$), $B_G(j,r)$ the ball of radius $r$ around $j$ in shortest-path distance $d_G$, $\lambda_r(A)=\max_u\sum_{v:\,d_G(u,v)>r}A_{uv}$ the largest attention any query sends beyond distance $r$, and $\Delta=\max_u(\max_v z_{uv}-\min_v z_{uv})$ the largest logit spread of any query.

\begin{proposition}[Three bounds on received attention]\label{prop:ceilings}
Under the setup above, for every node $j$ and every radius $r$,
\[
s_j(A)\;\leq\;\min\left\{\frac{1}{|\mathcal{O}(j)|},\;\;\lambda_r(A)+\bigl(1-\lambda_r(A)\bigr)\frac{|B_G(j,r)|}{n},\;\;\frac{e^{\Delta}}{e^{\Delta}+n-1}\right\}.
\]
In particular, on a vertex-transitive labelled input $s_j(A)=1/n$ for every node, independently of the learned parameters, and reaching received attention $\varepsilon$ requires $\Delta\geq\log\bigl((n-1)\varepsilon/(1-\varepsilon)\bigr)$.
\end{proposition}

The orbit term is the standard fact that equivariant maps are constant on orbits \citep{morris2024orbit}. What we add is that every bound can be computed head by head, so we can test whether any bound is tight enough to prevent the formation of a sink (Appendix~\ref{apd:details}). The symmetry bound is tight on symmetric graphs: every checkpoint with an equivariant positional encoding, even the dot-product models that sink heavily on molecules, spreads attention evenly across cycles of identical atoms and bonds ($C_8$ to $C_{24}$; Appendix~\ref{apd:details}). Symmetry alone prevents the sink, whatever the architecture. Conversely, Laplacian encodings, whose eigenvector signs and bases are arbitrary \citep{lim2023signnet}, break equivariance, and a dot-product Transformer then puts almost all attention ($0.999$) on one node.

On real molecules, neither graph-side bound is tight: atoms rarely have symmetric counterparts, attention is nonlocal (median $\lambda_1\geq0.91$), and the bounds rule out a sink in at most $\CERTmax\%$ of heads in any run. The third bound, on the spread of attention scores, does differ between architectures: GRIT clips its scores elementwise, so their spread never exceeds $5\lVert a\rVert_1$, whereas dot-product scores can grow without limit with the attended node's norm (Proposition~\ref{prop:range}). Yet this cap does not explain why GRIT rarely sinks: in our trained models, $5\lVert a\rVert_1$ ranges from $2.7$ to $11.7$ across layers, above the spread of $2.2$ a sink needs at ZINC size, and removing the clip leaves GRIT's sink rate low (Section~\ref{sec:scoring}). The graph thus leaves GRIT and GraphGPS equally free to sink on the same molecules, yet their sink rates differ more than four-fold: the models themselves must decide.

\section{The attention scoring function drives the gap}
\label{sec:scoring}

On identical ZINC graphs, GraphGPS and GRIT differ in heads, positional encoding, dropout, and message passing as well as in scoring, so we change one component at a time (Table~\ref{tab:dial}). Replacing only GRIT's logit with $z^{\mathrm{dot}}$ (labelled GRIT-dotlogit), leaving RRWP, the pair representation and the values unchanged, raises its sink rate from $\SRgrit$ to $\SRdotlogit$ (Figure~\ref{fig:headline}), above GraphGPS's. Removing message passing from GraphGPS does not lower its sink rate either ($\SRgpsnone$). Message passing is therefore not what suppresses sinks; the scoring function is.

The swap also slows training (test MAE $0.064$ to $0.373$), but two controls at matched accuracy show that this is not why it sinks. Without any graph structure (RRWP and bond types), GRIT's score and a dot product reach the same accuracy (test MAE $0.542$ and $0.543$, labelled GRIT-blind and GRIT-dot) yet sink at rates of $0.214$ and $0.532$. Likewise, GraphGPS with a virtual node matches GRIT's accuracy ($0.066$) yet sinks six times as often ($0.559$). Adding node RRWP and a bounded pair bias to GRIT-dot (labelled GRIT-dotbias) recovers over half the lost accuracy (test MAE $0.268$) but leaves its sink rate unchanged ($0.521$). Sink formation tracks mainly the form of the score; accuracy tracks access to graph structure. Even a loss that rewards concentrated attention lifts GRIT's sink rate only to $0.131$.

\begin{table}[htbp]
    \centering
    \caption{Changing each component individually on ZINC (mean $\pm$ $95\%$ CI; all arms in Table~\ref{tab:full}). \emph{Ablation ratio}: test MAE after removing all attention onto the most-attended node, divided by that for a random node.}
    \label{tab:dial}
    \small
    \let\oldpm\pm\renewcommand{\pm}{$\footnotesize${}\mathbin{\oldpm}{}}
    \begin{tabular}{lccc}
    \toprule
    Arm & Sink rate & Test MAE & Ablation ratio \\
    \midrule
    GRIT (gated scoring, RRWP) & $\SRgrit$ & $\MAEgrit$ & $\ABsrgrit$ \\
    GRIT-dotlogit (dot-product logit only) & $\SRdotlogit$ & $\MAEdotlogit$ & $\ABsrdotlogit$ \\
    GRIT-blind (gated scoring, no structure) & $\SRblind$ & $\MAEblind$ & $\ABsrblind$ \\
    GRIT-dot (dot-product scoring, no structure) & $0.532 \pm 0.061$ & $0.543 \pm 0.006$ & $\ABsrgritdot$ \\
    GRIT-dotbias (GRIT-dot + pair bias) & $\SRdotbias$ & $0.268 \pm 0.068$ & $\ABsrdotbias$ \\
    GraphGPS (GINE + dot attention) & $\SRgps$ & $\MAEgps$ & $\ABsrgps$ \\
    GraphGPS-none (MPNN removed) & $0.507 \pm 0.159$ & $0.200 \pm 0.017$ & $\ABsrgpsnone$ \\
    GraphGPS + virtual node & $0.559 \pm 0.177$ & $0.066 \pm 0.015$ & $\ABsrgpsvn$ \\
    GRIT + virtual node & $\SRgritvn$ & $\MAEgritvn$ & $\ABsrgritvn$ \\
    \bottomrule
    \end{tabular}
\end{table}

\textbf{Why large norms do not attract attention in GRIT.} A dot-product logit grows linearly with the attended node's hidden state, so a node with a large enough norm can claim almost all of a query's attention (Proposition~\ref{prop:range}). GRIT's score instead saturates: in layers 2 to 10, at least $\SATends\%$ of its coordinates sit at $0$ or $5$, so the logit is a sum of readout weights selected by which coordinates fire, not by input magnitude. Scaling up one node's hidden state confirms this (Table~\ref{tab:normint}): doubling it makes it a sink in up to $\NIhftwomaxdotlogit\%$ of heads in GRIT-dotlogit but at most $\NIhftwomaxgrit\%$ in GRIT, and even a $32$-fold increase reaches only $\NIhfmaxgrit\%$. Removing the clip (GRIT-noclamp) changes little: the model still sinks rarely ($0.113$), and a doubled norm makes a sink in at most $\NIhftwomaxnoclamp\%$ of heads, but an extreme $32\times$ norm increase does in up to $\NIhfmaxnoclamp\%$. GRIT thus grows high-norm nodes without attending to them.

\textbf{Giving the model a token to sink on.} A virtual node is the graph counterpart of <BOS>: chemically empty, connected to every node, akin to vision-transformer registers \citep{vision}, and exempt from the symmetry bound, since no automorphism can move it. On the symmetric cycles of Section~\ref{sec:theory}, where no real node may stand out, GraphGPS sends all of its attention to the virtual node, GRIT only about half to two-thirds on average, and inconsistently across seeds. On molecules, a virtual node leaves GRIT's sink rate low ($\SRgritvn$), even with a logit penalty toward it ($\SRgritvnbias$). GraphGPS with a virtual node usually sinks onto an atom rather than onto the virtual node, and the sink's value vector is about a tenth of the mean norm. Symmetry says where a sink may appear; the scoring function decides whether one does.

The ablation column of Table~\ref{tab:dial} shows what these sinks are worth: removing all attention to the most-attended node leaves the dot-product models' test MAE close to that for a random node ($\ABsrdotmin$ to $\ABsrdotmax$ times), but GRIT's twice as high ($2.06$). In dot-product models the sink is thus close to a no-op, as in LLMs \citep{barbero2025llms}, whereas GRIT's most-attended nodes matter to its predictions.

\section{Discussion}
\label{sec:discussion}

Graph Transformers based on GRIT largely do not form attention sinks on molecules, and the graph is not the reason: symmetry forbids sinks only on inputs far more regular than a molecule, and the attention we measure is too nonlocal for locality to matter. What matters most is the scoring function: GRIT's logit barely responds to a node's norm, and on molecules even a virtual node does not make it sink. The LLM sink therefore looks like a property of dot-product attention rather than of Transformers, consistent with sinks vanishing without softmax normalisation \citep{gu2024attention} or under an output gate \citep{qiu2025gated}. Read as attentional networks over a complete graph \citep{bronstein2021geometric}, Transformers differ from GRIT in two ways, a first token and a dot-product score, and our experiments separate the two: on molecules, dot-product scoring sinks with or without a designated token, while GRIT's gated scoring rarely sinks in either case. Whether sinks protect Graph Transformers against over-mixing remains open: our $10$-layer models show little over-smoothing to prevent (Appendix~\ref{apd:figures}).

Our study has limits: one gated-score architecture against dot-product baselines, on molecular graphs alone, at $10$ layers; several GRIT virtual-node quantities remain seed-dependent; and we measure reception rather than mixing, since a perfect sink can also be a complete mixer. Further architectures \citep[EGT, Graphormer, SAN;][]{yuan2025survey} and non-molecular graphs come next.

{
\footnotesize
\setlength{\bibsep}{5pt plus 0.3ex}
\bibliographystyle{plainnat}
\bibliography{references}
}

\section*{Acknowledgments}

I would like to thank Professor Michael Bronstein for his class on Geometric Deep Learning at the University of Oxford, where this research on Graph Transformers started as a course project.

The author acknowledges the use of resources provided by the Isambard-AI National AI Research Resource (AIRR)~\citep{mcintoshsmith2024isambardaileadershipclasssupercomputer}. Isambard-AI is operated by the University of Bristol and is funded by the UK Government’s Department for Business, Innovation, Science and Trade (DBIST) via UK Research and Innovation; and the Science and Technology Facilities Council [ST/AIRR/I-A-I/1023].

\appendix

\section{Proofs and supporting theory}\label{apd:theory}

\subsection{Proof of Proposition~\ref{prop:ceilings}}

\begin{proofsketch}
\emph{Orbit term.} For an automorphism $P$ of the labelled input, $PZ=Z$ and equivariance give $A(Z)=PA(Z)P^\top$, so with $s(A)=\tfrac1n A^\top\mathbf{1}$ and $P^\top\mathbf{1}=\mathbf{1}$ we get $Ps(A)=s(A)$: the sink score is constant on every orbit. Row-stochasticity gives $\sum_j s_j(A)=1$, so an orbit of size $m$ has $s_j(A)\leq1/m$, and a vertex-transitive input is one orbit of size $n$, forcing $s_j(A)=1/n$.

\emph{Locality term.} For $u\in B_G(j,r)$, row-stochasticity gives $A_{uj}\leq1$; for $u\notin B_G(j,r)$, $d_G(u,j)>r$ gives $A_{uj}\leq\sum_{v:\,d_G(u,v)>r}A_{uv}=\ell_u^{(r)}\leq\lambda_r(A)$. Therefore
\[
s_j(A)\;\leq\;\bigl(|B_G(j,r)|+\textstyle\sum_{u\notin B_G(j,r)}\ell_u^{(r)}\bigr)/n\;\leq\;\lambda_r(A)+(1-\lambda_r(A))\,|B_G(j,r)|/n,
\]
where the first inequality is the per-node form evaluated in the certificate of Appendix~\ref{apd:details}.

\emph{Range term.} Fix a query $u$. For $v\neq j$ the definition of $\Delta$ gives $z_{uv}\geq z_{uj}-\Delta$, so $A_{uj}=e^{z_{uj}}/\sum_v e^{z_{uv}}\leq e^{z_{uj}}/\bigl(e^{z_{uj}}+(n-1)e^{z_{uj}-\Delta}\bigr)=e^{\Delta}/(e^{\Delta}+n-1)$. Averaging over $u$ gives the same bound for $s_j(A)$, and $e^{\Delta}/(e^{\Delta}+n-1)\geq\varepsilon$ gives $\Delta\geq\log((n-1)\varepsilon/(1-\varepsilon))$. Taking the minimum of the three bounds proves the proposition.
\end{proofsketch}

\subsection{Score spread}

\begin{proposition}[Bounded and unbounded score spread]\label{prop:range}
For the GRIT score~\eqref{eq:score}, write $\widehat e_{uj}\in[0,5]^h$ for the clipped vector that $a$ reads out; hence $5\sum_{i:a_i<0}a_i\leq z_{uj}\leq5\sum_{i:a_i>0}a_i$ and $\Delta\leq5\lVert a\rVert_1$ whatever the node features, with the constant sharp over the box $[0,5]^h$. For dot-product scoring no such bound exists: if $x_j$ is the input state of node $j$ and $\langle q_u,W_Kx_j\rangle\neq0$ for some $u\neq j$, replacing $x_j$ by $\rho x_j$ makes $\Delta$ grow without bound in $\rho$, and a bounded edge-bias term cannot restore a bound.
\end{proposition}

\begin{proofsketch}
The clip and rectifier confine each coordinate to $[0,5]$, and a linear functional on that box is extremised at the two sign-selected corners, whose difference is $5\sum_i|a_i|$. For the dot product and $u\neq j$, only $k_j(\rho)=\rho W_Kx_j+b_K$ moves, so $z_{uj}$ is affine in $\rho$ with nonzero slope while the rest of row $u$ is fixed, and $\Delta$ is at least the gap between two entries of that row.
\end{proofsketch}

At $\varepsilon=0.3$ a sink needs a spread of $\DELTAzinc$ at $n=23$ and $\DELTApep$ at $n=269$, against $5\lVert a\rVert_1=\RANGEmin$ to $\RANGEmax$ across the ZINC layers, so the bound is loose (Section~\ref{sec:theory}).

\textbf{Reception is not mixing.} A sink score does not determine how much a layer mixes, the distinction that spectral and Dobrushin-type analyses of over-smoothing and attention propagation turn on \citep{wu2023demystifying,biccari2026rollout}. With $\kappa(A)=\tfrac12\max_{u,t}\lVert A_{u:}-A_{t:}\rVert_1$ the Dobrushin coefficient, $\operatorname{diam}(AY)\leq\kappa(A)\operatorname{diam}(Y)$ for any features $Y$. The identity $I_n$ and the uniform matrix $U_n=\tfrac1n\mathbf{1}\mathbf{1}^\top$ share the flat profile $s=\mathbf{1}/n$ yet have $\kappa(I_n)=1$ and $\kappa(U_n)=0$, while a perfect sink $\mathbf{1}e_v^\top$ has the maximal profile $s=e_v$ yet $\kappa=0$. We therefore measure mixing separately (Appendix~\ref{apd:figures}).

\section{Experimental details}\label{apd:details}

\begin{table}[ht]
\centering
\caption{Experimental hyperparameters. Peptides-func graphs are capped at 300 nodes to limit memory usage, and each quartile model is trained on 2{,}023 graphs of its quartile at batch size 16. All models are trained with AdamW, linear warmup, cosine annealing, and gradient clipping on a single GPU. Every other arm inherits its base model's configuration, changing only the named component.}
\label{tab:hyperparams}
\small
\begin{tabular}{llcccccccc}
\toprule
Model & Dataset & Dim & Heads & Layers & MPNN & PE & LR & Epochs & Batch \\
\midrule
GRIT & Peptides & 96 & 4 & 4 & -- & RRWP-17 & 3e-4 & 150 & 8 \\
GRIT & ZINC & 64 & 8 & 10 & -- & RRWP-21 & 1e-3 & 2000 & 32 \\
GraphGPS & ZINC & 64 & 4 & 10 & GINE & RWSE-28 & 1e-3 & 1500 & 32 \\
\bottomrule
\end{tabular}
\end{table}

\textbf{Arms.} GRIT-dotlogit replaces $z^{\mathrm{GRIT}}$ by $z^{\mathrm{dot}}$ in Eq.~\eqref{eq:score} and keeps everything else, including RRWP, the pair representation and the values; GRIT-noclamp removes the clip. Like the official code, GRIT-official clips the scalar logit to $[-5,5]$ rather than each coordinate; unlike that code, all our GRIT arms use one readout $a$ for every head. GRIT-blind removes node and pair RRWP and bond types; GRIT-dot is GRIT-blind with $z^{\mathrm{dot}}$ and no pair representation; GRIT-dotbias adds to GRIT-dot node RRWP and a bounded pair bias $a^\top\mathrm{clip}(We_{uj})$ in the logit. GRIT-entropy-1/2 add an attention-entropy penalty of two strengths. Virtual-node arms join one node to every atom, with the positional encoding padded or recomputed (PE rec.); GRIT + virtual node + bias adds a learned logit penalty toward it. GraphGPS-none removes the MPNN, GraphGPS-GatedGCN replaces it, and Transformer + LapPE is GraphGPS-none with sign-ambiguous Laplacian encodings (8 eigenvectors). Seeds are fixed per run, we report seed-controlled runs only, and checkpoints are best-validation.

\textbf{Interventions.} \emph{Sink ablation}: on 500 test graphs, the most-attended node (head-averaged received attention summed over layers) has its attention column zeroed in every layer, removing that mass (Table~\ref{tab:dial}) or renormalising the rows (Table~\ref{tab:full}); the control ablates a uniformly random node. \emph{Norm intervention}: at layer $\ell\in\{3,5,8\}$ the input state of one uniformly chosen real node is multiplied by $\rho$ before that layer's attention, on 200 test graphs, and we read that layer's attention only. \emph{Language models}: Pythia-70M/160M \citep{biderman2023pythia} and GPT-2 small \citep{radford2019gpt2} on 200 non-overlapping WikiText-2 test windows \citep{merity2017wikitext} of $T$ tokens without a BOS token, with the same sink-rate estimator.

\textbf{Certificate and probe.} For Proposition~\ref{prop:ceilings} we evaluate, on 300 test graphs per run, layer, and head, the row leakages $\ell_u^{(r)}$ and $\lambda_r(A)$ for $r\leq3$, the ball volumes, the orbits via VF2 with capped enumeration (a cap only shrinks orbits, so $1/|\mathcal{O}(j)|$ stays valid), and the per-node bound $U_j=\min\{1/|\mathcal{O}(j)|,\min_r(|B_G(j,r)|+\sum_{u\notin B_G(j,r)}\ell_u^{(r)})/n\}$. The locality bound was never violated, and within-orbit deviation was at most $\SPREADmax$ for every equivariant-PE run except GRIT-dotbias ($\SPREADdotbias$, with $\ORBVIOLdotbias$ orbit-bound violations in total), which doubles as an equivariance test. The probe passes cycles $C_8$--$C_{24}$ of identical atoms and bonds through the checkpoints; virtual-node arms use the training-time transform, so the virtual node is a singleton orbit beside the $n$ cycle nodes and the two are reported separately. On the cycle probe, received attention is uniform to $\PROBEdevmax$ in every equivariant-PE checkpoint except GRIT-dotbias and GRIT-dotlogit, which amplify float32 rounding (deviation up to $\PROBEdevdotlogit$) and are uniform to $\PROBEfpdot$ in float64. Real nodes of virtual-node arms stay uniform to $\PROBEdevvn$.

\section{Additional results}\label{apd:figures}

\textbf{Training verification.} GRIT reaches test MAE $0.064$ on ZINC, close to the published result \citep{GRIT}, and test AP $0.634$ on the full Peptides-func dataset ($0.612$ on the smallest quartile, $0.499$ on the largest), below the published result; GraphGPS reaches $0.072$ on ZINC. At GRIT's 2{,}000-epoch budget, GRIT-dotlogit appears under-trained, with an unconverged training loss.

\textbf{Sink ablations.} Table~\ref{tab:dial} removes the attention onto the most-attended node outright. If each row is instead renormalised, so that this attention is spread over the other nodes, GRIT's ablation ratio falls from $2.06$ to $1.35$, but every gated arm ($1.25$ to $1.61$) stays above every dot-product arm ($1.00$ to $1.23$) except GraphGPS-none, whose ratio varies widely across seeds ($2.03$; Table~\ref{tab:full}).

\textbf{Thresholds and size.} GRIT sinks less than each of the seven dot-product arms we swept at every threshold from $\varepsilon=0.1$ to $0.8$ (closest at $\varepsilon=0.8$: $0.02$ vs.\ $0.05$), and the threshold-free peak share $n\max_j s_j$ separates gated arms ($2.5$ to $3.6$) from dot-product arms ($6.0$ to $13.1$) with no overlap (Table~\ref{tab:full}). On Peptides-func, GRIT's largest sink score falls with graph size ($0.135$ to $0.066$) while its peak share rises ($7.4$ to $16.1$), as a score spread that does not grow with $n$ predicts.

\textbf{Language models.} With the same estimator, the three language models reach sink rates of \LMsrA\ at $T=23$ and \LMsrB\ at $T=269$. Compared to our experiments, these language models were trained on far longer contexts, which bring stronger sinks \citep{barbero2025llms}.

\textbf{Virtual-node details.} At the peak-sinking layer, the virtual node of GRIT + virtual node is never the most-attended node, and it almost never receives enough attention to be a sink. In GraphGPS + virtual node it is the most-attended node in $10\%$ of graphs; when it is, its value norm is about a tenth of the mean.

\textbf{Over-mixing.} From first to last layer the Dirichlet energy relative to the squared feature norm grows in every arm (on average $\RAYminrest$- to $\RAYmaxrest$-fold) except GRIT + virtual node (PE rec.), where it falls to \RAYvnpe\ of its input value, and the distance from consensus $\lVert X-\mathbf{1}\bar x^\top\rVert_F/\lVert X\rVert_F$ rises in every arm: at $10$ layers there is little over-smoothing for a sink to prevent.

\begin{table}[ht]
\centering
\caption{All ZINC arms (mean $\pm$ $95\%$ CI over three seeds, five for GraphGPS). $n\max_j s_j$: attention received by the most-attended node, as a multiple of the uniform share. \emph{Ablation ratio}: test MAE after removing all attention onto the most-attended node, divided by that for a random node.}
\label{tab:full}
\footnotesize
\setlength{\tabcolsep}{4pt}
\begin{tabular}{lccccc}
\toprule
Arm & Sink rate & Test MAE & $n\max_j s_j$ & Abl.\ removed & Abl.\ renorm. \\
\midrule
GRIT & $\SRgrit$ & $\MAEgrit$ & $\NSgrit$ & $\ABsrgrit$ & $\ABRsrgrit$ \\
GRIT-dotlogit & $\SRdotlogit$ & $\MAEdotlogit$ & $\NSdotlogit$ & $\ABsrdotlogit$ & $\ABRsrdotlogit$ \\
GRIT-noclamp & $\SRnoclamp$ & $\MAEnoclamp$ & $\NSnoclamp$ & $\ABsrnoclamp$ & $\ABRsrnoclamp$ \\
GRIT-official & $\SRofficial$ & $\MAEofficial$ & $\NSofficial$ & $\ABsrofficial$ & $\ABRsrofficial$ \\
GRIT-blind & $\SRblind$ & $\MAEblind$ & $\NSblind$ & $\ABsrblind$ & $\ABRsrblind$ \\
GRIT-dot & $\SRgritdot$ & $\MAEgritdot$ & $\NSgritdot$ & $\ABsrgritdot$ & $\ABRsrgritdot$ \\
GRIT-dotbias & $\SRdotbias$ & $\MAEdotbias$ & $\NSdotbias$ & $\ABsrdotbias$ & $\ABRsrdotbias$ \\
GRIT-entropy-1 & $\SRentone$ & $\MAEentone$ & $\NSentone$ & $\ABsrentone$ & $\ABRsrentone$ \\
GRIT-entropy-2 & $\SRenttwo$ & $\MAEenttwo$ & $\NSenttwo$ & $\ABsrenttwo$ & $\ABRsrenttwo$ \\
GRIT + virtual node & $\SRgritvn$ & $\MAEgritvn$ & $\NSgritvn$ & $\ABsrgritvn$ & $\ABRsrgritvn$ \\
GRIT + virtual node + bias & $\SRgritvnbias$ & $\MAEgritvnbias$ & $\NSgritvnbias$ & $\ABsrgritvnbias$ & $\ABRsrgritvnbias$ \\
GRIT + virtual node (PE rec.) & $\SRgritvnpe$ & $\MAEgritvnpe$ & $\NSgritvnpe$ & $\ABsrgritvnpe$ & $\ABRsrgritvnpe$ \\
GraphGPS (GINE) & $\SRgps$ & $\MAEgps$ & $\NSgps$ & $\ABsrgps$ & $\ABRsrgps$ \\
GraphGPS-GatedGCN & $\SRgpsgated$ & $\MAEgpsgated$ & $\NSgpsgated$ & $\ABsrgpsgated$ & $\ABRsrgpsgated$ \\
GraphGPS-none & $\SRgpsnone$ & $\MAEgpsnone$ & $\NSgpsnone$ & $\ABsrgpsnone$ & $\ABRsrgpsnone$ \\
Transformer + LapPE & $\SRlappe$ & $\MAElappe$ & $\NSlappe$ & $\ABsrlappe$ & $\ABRsrlappe$ \\
GraphGPS + virtual node & $\SRgpsvn$ & $\MAEgpsvn$ & $\NSgpsvn$ & $\ABsrgpsvn$ & $\ABRsrgpsvn$ \\
GraphGPS + virtual node (PE rec.) & $\SRgpsvnpe$ & $\MAEgpsvnpe$ & $\NSgpsvnpe$ & $\ABsrgpsvnpe$ & $\ABRsrgpsvnpe$ \\
\bottomrule
\end{tabular}
\end{table}

\begin{table}[ht]
\centering
\caption{\% of heads in which one node, whose input is scaled by $\rho$, becomes a sink (max over layers).}
\label{tab:normint}
\footnotesize
\begin{tabular}{lcccccc}
\toprule
 & GRIT & GRIT-noclamp & GRIT-official & GRIT-dotlogit & GRIT-dot & GraphGPS \\
\midrule
$\rho=2$ & $\NIhftwomaxgrit$ & $\NIhftwomaxnoclamp$ & $\NIhftwomaxofficial$ & $\NIhftwomaxdotlogit$ & $\NIhftwomaxgritdot$ & $\NIhftwomaxgps$ \\
$\rho=8$ & $\NIhfeightmaxgrit$ & $\NIhfeightmaxnoclamp$ & $\NIhfeightmaxofficial$ & $\NIhfeightmaxdotlogit$ & $\NIhfeightmaxgritdot$ & $\NIhfeightmaxgps$ \\
$\rho=32$ & $\NIhfmaxgrit$ & $\NIhfmaxnoclamp$ & $\NIhfmaxofficial$ & $\NIhfmaxdotlogit$ & $\NIhfmaxgritdot$ & $\NIhfmaxgps$ \\
\bottomrule
\end{tabular}
\end{table}
\end{document}